\documentclass[10pt,twocolumn,letterpaper]{article}
\usepackage{cvpr}              

\usepackage{graphicx}
\usepackage{amsmath}
\usepackage{amssymb}
\usepackage{booktabs}
\usepackage{float}
\usepackage[pagebackref,breaklinks,colorlinks]{hyperref}
\usepackage[nodisplayskipstretch]{setspace}
\usepackage[capitalize]{cleveref}
\crefname{section}{Sec.}{Secs.}
\Crefname{section}{Section}{Sections}
\Crefname{table}{Table}{Tables}
\crefname{table}{Tab.}{Tabs.}

\def\cvprPaperID{*****} 
\def\confName{CVPR}
\def\confYear{2022}

\begin{document}

\title{Predict NAS Multi-Task by Stacking Ensemble Models using GP-NAS}

\author{Ke Zhang\\
{\tt\small kezzhang@gmail.com}
}
\maketitle

\begin{abstract}
   Accurately predicting the performance of architecture with small sample training is an important but not easy task. How to analysis and train data set to overcome over fitting is the core problem we should deal with. Meanwhile if there is the multi-task problem, we should also think about if we can take advantage of their correlation and estimate as fast as we can. In this track, Super Network builds a search space based on ViT-Base. The search space contain depth, num-heads, mpl-ratio and embed-dim. What we done firstly are preprocessing the data based on our understanding of this problem which can reduce complexity of problem  and probability of over fitting. Then we tried different kind of models and different way to combine them. Finally we choose stacking ensemble models using GP-NAS with cross validation. Our stacking model ranked 1st in CVPR 2022 Track 2 Challenge.
\end{abstract}

\section{Introduction}

As limited data sources, the cost of collection high precision large-scale data set, the difficult of labeling data manually, limitation of computing power, the requirement of fast estimation, training based on small size and predict accurately is more and more important in widely different industry such as computer vision, finance, E-commerce, medicine and so on. Meanwhile, as we have more and more different source to collect data, we should consider how to handle multi-task simultaneously without sacrificing precision or take advantage of their collaboration.

Neural Architecture Search (NAS)\cite{elsken2019neural}\cite{real2019regularized}\cite{zoph2018learning} has been used to design deep neural network architectures successfully. It outperform hand-designed models in many computer vision tasks. However, there are still several issues need to be solved when we use NAS kind of algorithm. For example, inconsistent with the performance of the same network trained independently. Poor performance in small sample sized training and prediction. Recently GP-NAS(Gaussian Process based Neural Architecture Search) was introduced\cite{li2020gp}, the correlation are modeled by the kernel function and mean function. Bayesian theorem and information theory was implemented deeply in this model which theoretically ensure the good performance especially when sample size is small. 

However, we find that naively and simply throw data into GP-NAS can not give us relatively good result. We think there are several reasons. First of all, this kind of model is sensitive to the input which require accurately preprocessing data. Secondly, purely running this model can not fully mining most of information from data especially when there are nonlinear or complex feature inside the data set. 

In this paper, after we deeply analysis this problem and try different kind of model, we finally choose to stack several ensemble models by GP-NAS which give us quite good performance in CVPR 2022 Track2 Challenge. 

\section{How we deal this problem}
\subsection{Raw data}
In this challenge, Super Network builds a search space based on ViT-Base for 8 tasks. The search space involves the number of depth, the number of heads, the mlp-ratio and embed-dim. For each task, the training data include 500 samples whose input is the model structure and whose label is the relative ranking of structure. While the test data includes 99500 samples. We can find training sample size is relatively small compare with test data which required the ability of accurately fitting based on small sample data set. Obviously this is the purpose of this track challenge that predicting the performance of any architecture accurately with small or without training.

\subsection{Data preprocessing}

For this problem, most straight forward and direct way to do is training all the network using default structure value. The problem of this way is singularity (embed-dim doesn't contain information) and does not take advantage of background information. Therefore, we pre process the data by following steps:
\begin{itemize}
\item Transfer the depth encoding(j,k,l) to integer. Here we consider ordinal encoding instead of one-hot encoding as we assume depth has some monotonic correlation with predictability, that deeper depth with more information. the experiment also confirm our thought.\cite{potdar2017comparative}

Ordinal encoding: 
\begin{equation}
  (j,k,l) \Rightarrow (1,2,3)
  \label{eq:e1}
\end{equation}

One-Hot encoding:
\begin{equation}       
  (j,k,l) \Rightarrow
\left(   
  \begin{array}{ccc}   
    1 & 0 & 0\\  
    0 & 1 & 0\\  
    0 & 0 & 1\\ 
  \end{array}
\right)
\label{eq:e2}
\end{equation}

\item Getting rid of embed-dim feature parameters as it doesn't include any information which avoid singularity and reduce the dimension of problem by 1/4. On the other hand when the actual depth of a sub-network is less than 12, its encoding trailing end is padded with 0, here we change 0 to 2 as we assume 2 represent neutral information. After this transformation, depth encoding feature has average 0.015 correlation with last four encoding features instead of average 0.75. Finally we normalize data from (1,2,3) to (-1,0,1).

\item Using inverse sigmoid function as activation function\cite{choong2017evaluation}. As we known, ranking data follow uniform distribution. By inverse sigmoid function we transfer uniform distribution to Gaussian distribution which is best choice for most model. After we get prediction, we use sigmoid function to transfer output back to uniform distribution as rank output, then we round up to nearest integer.

\begin{figure}[H]
  \centering
  \includegraphics[width=1.\linewidth]{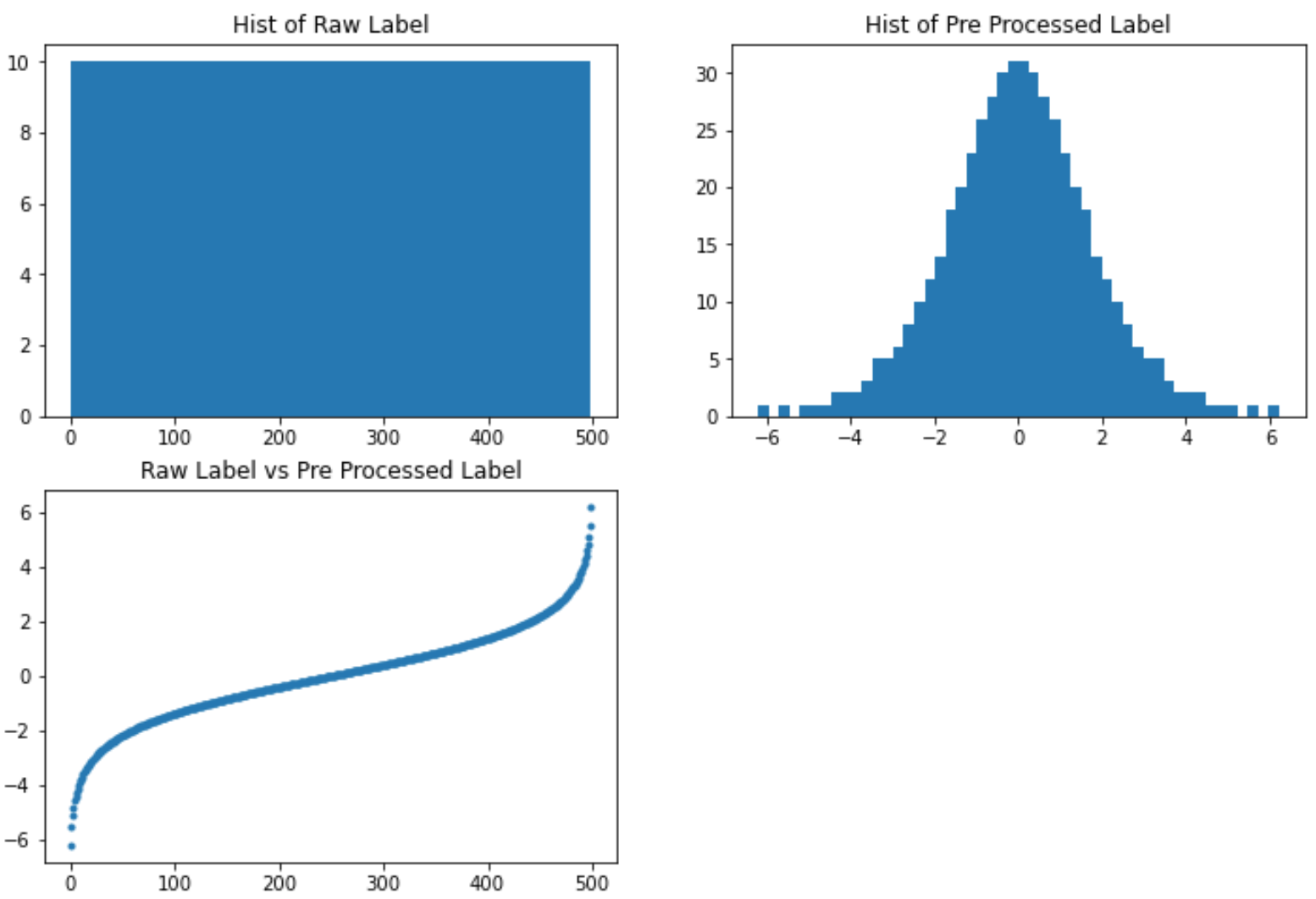}
   \caption{Using inverse sigmoid function as activation function to transfer rank label (Uniform distribution) to new label (Gaussian distribution)}
  \label{fig:Prep_label}
\end{figure}
The above figures shows the histogram of raw label, histogram of preprocessing label and relationship between raw label and preprocessing label.

\begin{equation}
  z = \frac{\log(y+1)}{n-y},y \in [0,n-1]
\label{eq:e3}
\end{equation}
\cref{eq:e3} is the inverse sigmoid function we mentioned above. 

\item we also try lots of other ideas such as different activation function, adding Gaussian noise and so on. None of them obviously improve across different task.
\end{itemize}
\subsection{Model Construction}
After we preprocessed our raw data set, then we consider to construct our training and testing framework. we tried lots of models and different combined ways (details are in next section), finally we choose to stack the ensemble models\cite{wolpert1992stacked}\cite{hastie2009elements}\cite{ghojogh2019theory} with final estimator GP-NAS\cite{dvzeroski2004combining}.

There are several steps to get our final result. Firstly, we split our data set into training data set and testing data set. We train our sub model by full training data set. Then we split our training data set to K fold and use K-1 to get prediction from each sub model\cite{kohavi1995study}. Secondly, for each K, we use prediction from each sub model as training data set and remain 1 fold as validated data to train GP-NAS and repeat K times. Now we can predict based on GP-NAS to get our second layer prediction. 

Finally, we post processing our data set that using sigmoid function as activation function to transfer our second layer prediction to uniform distribution and round it up to nearest integer as the rank prediction.

\begin{equation}
  s=\frac{n-1}{1+\exp(-z)}, z\in[-\infty,+\infty]
\label{eq:e5}
\end{equation}
\cref{eq:e5} is the sigmoid function we used. 
\begin{figure}[H]
  \centering
  \includegraphics[width=1.\linewidth]{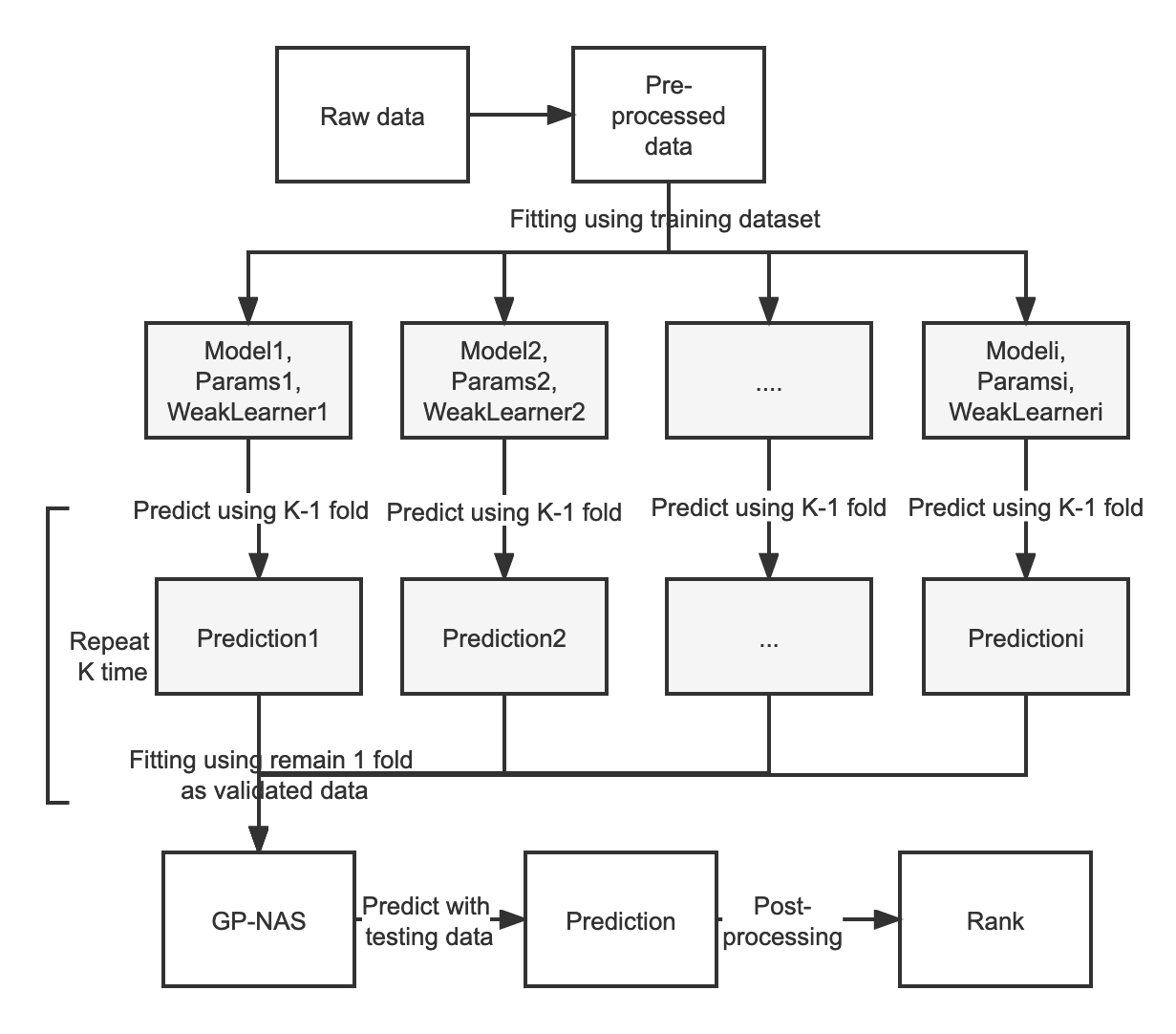}

   \caption{End to end training and predicting process}
   \label{fig:a5}
\end{figure}
Until now, we almost constructed our model framework. However, this doesn't mean we already get the good result from this framework. At beginning we use same parameters such as learning rate, loss function and depths of model for different tasks,but the output is not as good as final result. This phenomenon definitely make sense because different data set have different signal noise ratio and property. For example, higher signal noise ratio task require larger learning rate, more non-linear and complexity structure need larger maximum of depth. In a word, depending on the problem, we should find a different variance and bias trade off to predict well but avoid over fitting.

\subsection{Technical discussions and improvements}
Previous section we introduce our framework to deal with this task. Before that we tried different models and ways to combine. Now let's look at what we tried and how we get our final model.
\begin{itemize}

\item Firstly, baseline code was provided to us. With only GP-NAS model\cite{li2020gp}, the final score is about 0.67 (all the score we calculated is Kendall-tau).

\item Besides, we considered other machine learning models such as linear regression, random forest model, kernel ridge regression, tree model,light Gradient Boosting\cite{ke2017lightgbm}, XGBoost\cite{chen2016xgboost},Cat Gradient Boosting\cite{dorogush2018catboost} and so on. We find Gradient Boosting or other Boosting based algorithm\cite{friedman2001greedy} have best average result. When we firstly try it, we get score about 0.73. after tuning parameters, we get 0.78. This experiment let us narrowed our sub-model pool inside boosting based algorithm.

Boosting based algorithm is a tree-based ensemble model that ensemble of several weak learners which usually are decision trees. 
Decision trees are relatively weak on their own usually solely on yes/no questions. For boosting based algorithm, Each tree’s goal is not to maximize its own utility function on a randomly delegated subset of data, but to fit on residual of previous tree which performances relatively well in non-linear and complex problem.

Lots of experiment showed boosting based algorithm has these advantages compare with others: usually provides predictive accuracy that hardly be beaten. Very flexible that can optimize on different loss functions and provides several hyper parameter tuning options. Most important, no requirement of data preprocessing, not sensitive to distribution of data and missing point.

\item As our problem is multi-task, could we train different task at same time? We consider Multivariate Gradient Boosting based on MultiRMSE\cite{dorogush2018catboost} from package catboost. 
\begin{equation}
  MultiRMSE = \sqrt{\frac{\sum_{i=1}^{N}\sum_{d=1}^{dim}(a_{i,d}-t_{i,d})^2w_{i}}{\sum_{i=1}^{N}w_{i}}}
  \label{eq:e7}
\end{equation}
However, the result is even worse than univariate gradient boosting method. The reason maybe as follow: different task is so different that should consider different parameters; there are no obvious lead or lag relationship between different task; we can't reduce the noise in this multivariate Gradient Boosting algorithm effectively.

\item As our training sample size is so small, in order to avoid over fitting, we try to ensemble our models. We choose GBRT\cite{friedman2001greedy},HISTGB\cite{ke2017lightgbm},CATGB\cite{dorogush2018catboost},XGBoost\cite{chen2016xgboost},lightGB\cite{ke2017lightgbm} as our sub models. The obvious way is averaging all the result, this result improved to about 0.79.

\item After naive averaging, then we think about stacking gradient boosting group of model by a final estimator. After several experiment and comparison. We pick GP-NAS as our final regressor. Stacking is a general procedure where final estimator is trained to combine the individual learners. Stacking allows us to use the strength of each individual estimator by using their output as input of a final estimator. In order to avoid over fitting, we use full training data set only in sub model training, but training our final model with cross validation. We use K-1 fold to predict sub-model and remain 1 fold as validate data and repeat for each K.  As scikit-learn already provide us this regressor\cite{wolpert1992stacked}, we modified GP-NAS\cite{li2020gp} as a scikit-learn API. Now we get result about 0.792.

\item Until now our stack model still use same parameters for different task such as loss function, learning rate and depth of model. Then we consider select different loss function, learning rate and depth for different task as we assume different data set and task have different signal noise ratio and different features. Meanwhile we add two more GBRT, CATGB sub-model with different loss function( Huber or MSE) which also give us improvement. This round we get result about 0.798.

\begin{figure}[H]
  \centering
  \includegraphics[width=1.\linewidth]{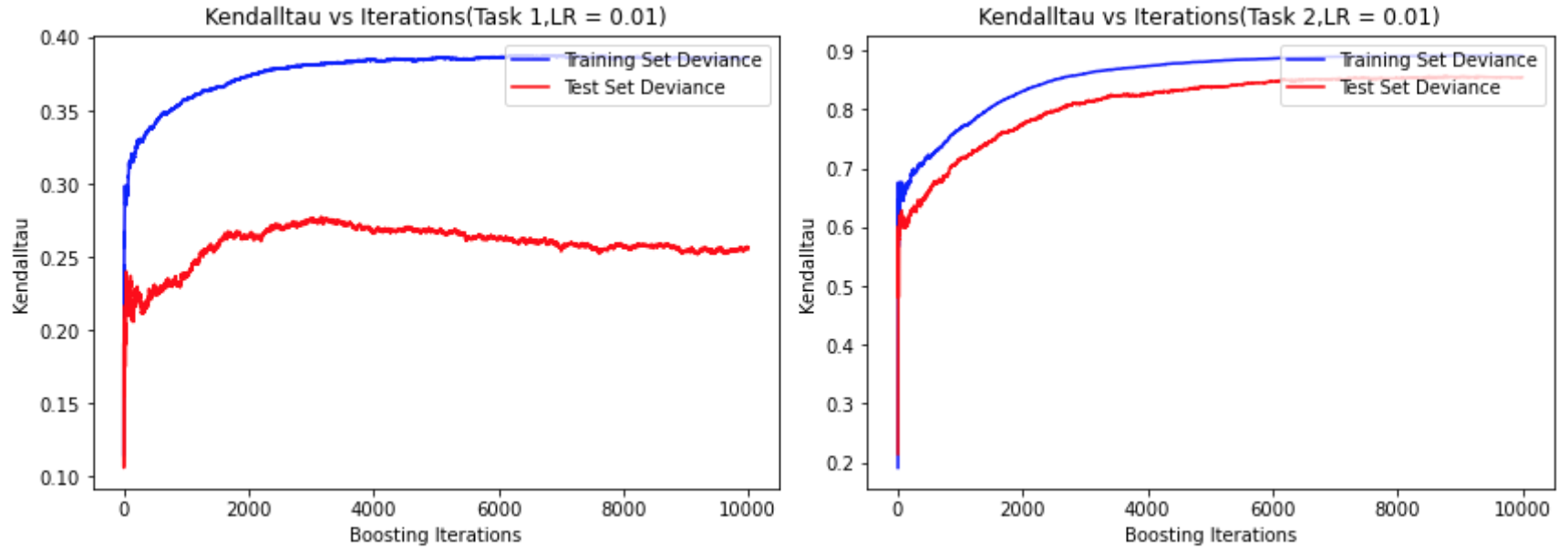}
   \caption{Task 1 and Task 2 Kendall-tau with different boosting iter}
   \label{fig:a6}
\end{figure}

\cref{fig:a6} shows different task has total different signal to noise ratio and feature even with same model. In this example, we use same parameter and model for Task 1 and Task 2 which show totally different phenomenon.

\item Also we modified and tuning our final estimator GP-NAS, such as ridge parameter and estimator prior. Instead of identity matrix, we choose inv(X.T*X) as our prior covariance matrix like equation (9). We get final score 0.7991 in leader-board A and 0.79849 in leader-board B. 
\end{itemize}

\begin{equation}
  Mean(\beta|X,Y) = {(X^TX)}^{-1}{X^TY}
  \label{eq:e8}
\end{equation}

\begin{equation}
  Var(\beta|X,Y) = {\sigma^2}{(X^TX)}^{-1}
  \label{eq:e9}
\end{equation}
\cref{eq:e8}\cref{eq:e9} are mean prior and variance prior of estimator.

\begin{table}[H]
  \centering
  \begin{tabular}{@{}lc@{}}
    \toprule
    Model & Kendall-tau \\
    \midrule
    GP-NAS Baseline & 0.67 \\
    Gradient Boosting & 0.78 \\
    Multivarate Gradient Boosting & 0.787 \\
    Averaging 5 ensemble algorithm & 0.79\\
    Stacked 5 algo with same params of tasks & 0.792\\
    Stacked 7 algo with diff params of tasks & 0.798\\
    Stacked 7 algo with inverse cov as prior & 0.7991\\
    \bottomrule
  \end{tabular}
  \caption{Result comparsion}
  \label{tab:t1}
\end{table}
\cref{tab:t1} show the model we tried and their Kendall-tau score.

\section{Conclusion}
In this paper, we explain how we preprocessing raw data, the model we tried, and the detail of last version of model framework. It shows stacking several ensemble models with GP-NAS give us a quite great result. As we consider cross validation during fitting, the prediction result by testing data is also considerably good. As time limitation, there are lots of idea we haven't deeply go through, such as improving the multivariate Gradient Boosting, stacking multivariate ensemble models, different kernel and prior in GP-NAS which maybe consider in the future. 
{\small
\bibliographystyle{ieee_fullname}
\bibliography{egbib}
}

\end{document}